\documentclass{article}

\usepackage{arxiv}

\usepackage[utf8]{inputenc}
\usepackage[T1]{fontenc}
\usepackage{hyperref}
\usepackage{url}
\usepackage{booktabs}
\usepackage{amsfonts}
\usepackage{nicefrac}
\usepackage{microtype}
\usepackage{graphicx}
\usepackage{amsmath,amssymb}
\usepackage{mathtools}
\usepackage{bm}
\usepackage{mathrsfs}
\usepackage{amsthm}
\usepackage{multirow}
\usepackage{float}
\usepackage{xcolor}
\usepackage{array}
\usepackage{enumitem}
\usepackage{caption}

\theoremstyle{plain}

\theoremstyle{definition}

\newcommand{\R}{\mathbb{R}}

\newcommand{\Heis}{\mathrm{Heis}_3}

\title{A Heisenberg Lift Descriptor for Order Sensitive Online Handwriting Recognition}

\author{
  Hassan Ugail \\
  Centre for Visual Computing and Intelligent Systems \\
  University of Bradford \\
  Bradford, United Kingdom \\
  \\
  \And
  Newton Howard \\
  School of Individualized Study \\
  Rochester Institute of Technology \\
  New York, United States \\
}

\begin{document}
\maketitle

%-----------------------------------------------------------------------
\begin{abstract}
Online handwriting recognition systems typically represent pen
trajectories through fixed-length Euclidean shape descriptors that
capture the spatial outline of each stroke, but are insensitive to the
order in which that outline is produced.
Two strokes that trace the same region of the plane in opposite
directions are indistinguishable to any such order-blind representation,
yet their traversal directions may carry decisive class information in
characters where loop orientation and stroke sequencing matter.
This paper introduces a Heisenberg-lift framework that addresses this
gap through a compact, interpretable, order-sensitive augmentation for
online pen-trajectory features.
The simplest instance is the terminal signed area, a single
parameter-free scalar appended to an existing Euclidean descriptor at
negligible computational cost.
Evaluated on two standard online handwriting benchmarks, this one-scalar
addition, consistently raises classifier accuracy over the Euclidean
baseline.
On the hardest character pair in our study, the letters o and y, the
signed area alone achieves perfect separation while the Euclidean
baseline falls short.
The advantage grows further under additive coordinate noise, a
practically relevant degradation in pen-trajectory data.
A richer fifteen-dimensional extension, derived from a noncommutative
Heisenberg-group subdivision scheme, provides additional gains in noisy
and loop-structured conditions.
Dimension-matched statistical controls confirm that all improvements
reflect geometric information rather than feature-count inflation.
The resulting descriptor is lightweight, closed-form, and directly
interpretable, making it a practical augmentation for online handwriting
and related document-trajectory classification pipelines in which the
direction of stroke execution carries discriminative information.
\end{abstract}


%%=========================================================%%
\section{Introduction}
\label{sec:intro}
%%=========================================================%%

Online handwriting recognition is a core document-analysis task in which
a classifier must assign a class label to a two-dimensional pen
trajectory recorded as an ordered sequence of positions.
The dominant approach to feature engineering for such trajectories
computes fixed-length Euclidean descriptors, including arc length,
bounding-box dimensions, curvature statistics, and discrete Fourier
amplitudes, and passes them to a conventional pattern classifier
\cite{zhang2004,ghosh2022,shanthi2010}.
This strategy is computationally inexpensive, straightforward to
implement, and frequently effective.
It captures the overall spatial shape of a pen stroke well.
What it does not capture, however, is the order in which that shape was
produced.

This omission matters more than it might first appear.
A pen trajectory is not merely a static geometric object but also a
path traced through time by a moving instrument.
Two strokes may occupy almost exactly the same region of the plane and
yet differ in the direction in which a loop is closed, the order in
which segments are drawn, or the way in which the pen returns towards
its starting point.
Any descriptor that is invariant under monotone reparametrisation is
blind to that distinction.
In other words, if only the spatial outline is encoded, a trajectory and
its time-reversal become indistinguishable.
No amount of additional Euclidean shape information can recover what was
never represented in the first place.

In online handwriting recognition, this limitation has direct practical
consequences for document-analysis systems.
Many characters are distinguished not only by their visible form but
also by the manner in which they are written.
Loop orientation, stroke direction, and stroke sequencing all carry
class information that Euclidean shape statistics discard entirely
\cite{bhattacharya2012,toselli2019}.
Earlier work on handwritten numeral verification has already established
that compact descriptors achieve their best results when they encode the
right geometric signal rather than simply the longest possible feature
vector \cite{zhang2004}.
The core difficulty is therefore to represent the missing temporal
information, not to extend a descriptor that already lacks it.
This gap is illustrated concretely in the UCI Character Trajectories
dataset, which constitutes one of the two handwriting benchmarks
evaluated in the present study.
In the hardest character pair identified in our experiments, the letters
\texttt{o} and \texttt{y}, the Euclidean descriptor, level-two path
signatures, and dimension-matched random augmentations all plateau at
the same accuracy ceiling.
Appending further random dimensions produces no improvement.
The limitation is therefore informational rather than dimensional, and
cannot be remedied by scaling up a representation that encodes only
spatial shape.
By contrast, the terminal signed area alone separates this character
pair perfectly, which demonstrates that traversal order is the decisive
discriminative signal in this specific case.

The starting point of the present work is the observation that the
signed area accumulated along a planar curve provides a natural way of
encoding traversal order.
Unlike Euclidean shape statistics, signed area changes sign when the
direction of traversal is reversed.
At the same time, it remains tied to the geometry of the curve in a
clear and interpretable way.
This makes it a particularly appealing candidate for a lightweight
order-sensitive descriptor.
The appropriate algebraic setting for this quantity is the
three-dimensional Heisenberg group
$\Heis = (\R^3, \cdot)$, whose multiplication law contains precisely the
antisymmetric cross-term associated with signed area.
When a pre-smoothed planar trajectory is lifted into this group, the
resulting third coordinate records the cumulative signed-area
accumulation along the path.
Rather than treating the signed area only as a terminal scalar, this
construction allows its dynamics to be tracked throughout the entire
trajectory.

That idea leads naturally to a layered feature design.
At its most basic, the terminal signed area $z(T)$ can simply be added
to an existing Euclidean descriptor as a single extra scalar.
This is the minimal form of the method and, as the experiments will
show, it already recovers most of the available gain on clean
multi-class data.
A fuller extension is obtained by refining the lifted trajectory with a
noncommutative interpolatory subdivision scheme $S_H$, adapted from the
classical four-point scheme of Dyn, Gregory and Levin
\cite{dyn1987,dyn1992} and analysed as a geometric object in its own
right in a companion study of Heisenberg-group
subdivision~\cite{ugail2026subdiv}.
This refinement produces a stable, dense profile of signed-area
accumulation from which a compact set of temporal summary statistics can
be extracted.
The richer profile is not intended to replace the minimal scalar
augmentation, but to extend it in the settings where temporal structure
matters more strongly, especially under noise and for short or
loop-heavy control polygons.

\begin{figure}[t]
\centering
\includegraphics[width=\linewidth]{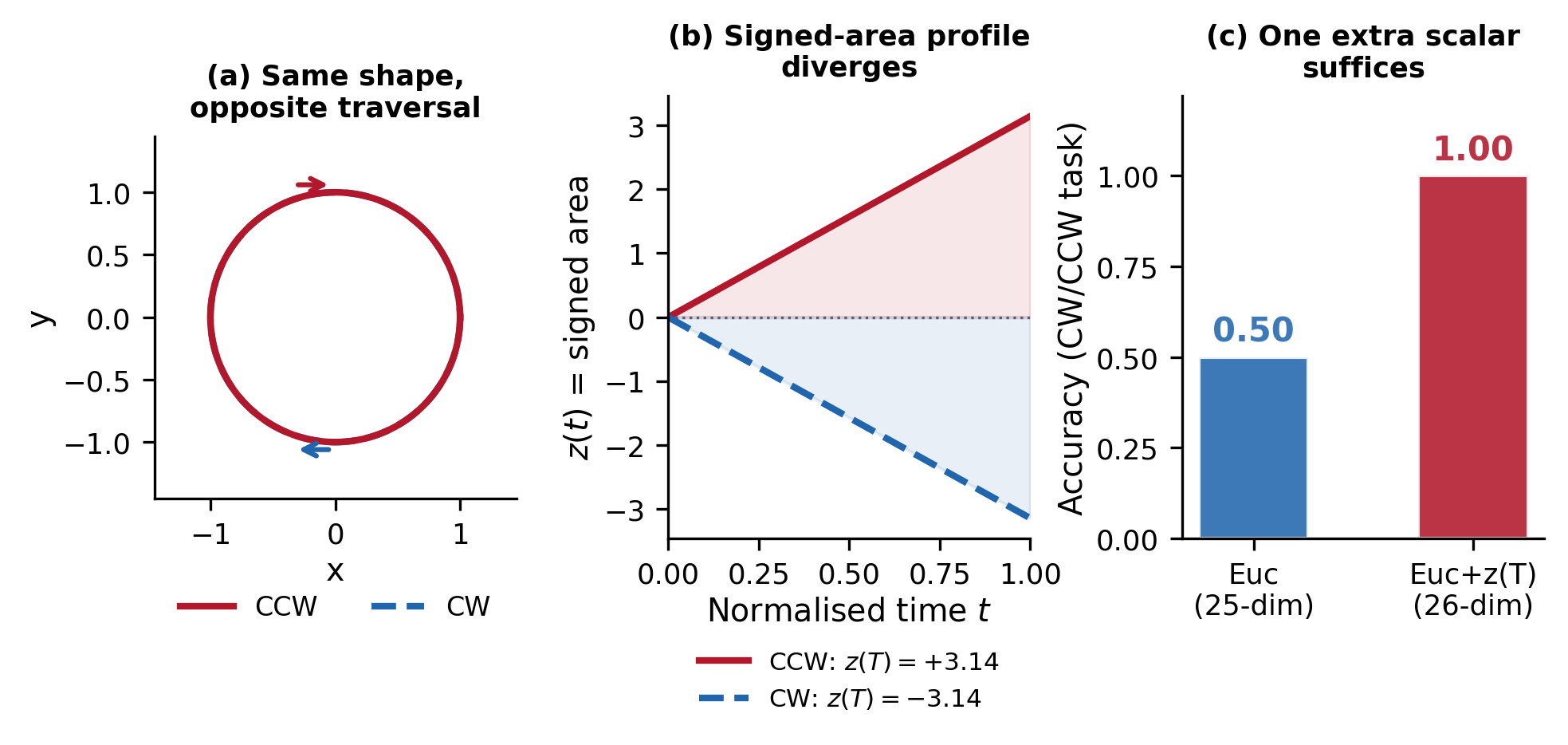}
\caption{The order-blindness problem.
  Panel~(a): a counter-clockwise (CCW) and a clockwise (CW) loop of the
  same radius are spatially identical, meaning all Euclidean shape statistics
  are equal.
  Panel~(b): the signed-area accumulation $z(t)$ is the sole
  distinguishing signal, growing positive for CCW ($z(T)=+3.14$)
  and negative for CW ($z(T)=-3.14$).
  Panel~(c): adding $z(T)$ to a 25-dimensional Euclidean baseline
  raises accuracy from 0.50 to 1.00 on the synthetic CW/CCW task.}
\label{fig:motivation}
\end{figure}

The aim of this paper is not to propose a general-purpose replacement
for strong sequence classifiers, nor to claim the introduction of a new
mathematical invariant.
The signed area accumulated along a planar curve is already well
understood, particularly within the path-signature literature where it
appears as the antisymmetric level-two term.
The contribution is to demonstrate that this quantity can be packaged
into a compact, interpretable, and practically useful feature
augmentation specifically for online handwriting and related
pen-trajectory classification tasks in document analysis.
The empirical study is organised around that claim, drawing exclusively
on established online handwriting benchmarks and reporting results under
both clean and noise-degraded conditions of the kind that arise in
realistic document-analysis settings.

The contributions of the paper are best understood in three tiers.

\textit{Tier~1 (minimal scalar augmentation).}
We show that appending $z(T)$, without learned parameters, to a
25-dimensional Euclidean descriptor yields Euc+$z(T)$ (26-dim), which is
significantly more accurate than Euclidean alone on two independent
benchmarks.
On the hardest character pair, \texttt{o} versus \texttt{y}, the scalar
$z(T)$ alone achieves perfect separation while 25 Euclidean features
reach only $0.964$, confirming that the gain is informational rather
than dimensional.

\textit{Tier~2 (Heisenberg temporal profile).}
We construct a 15-dimensional profile of $z(t)$ dynamics by combining
the Heisenberg lift with $S_H$ subdivision.
This fuller descriptor adds $+1.3$\,pp on clean UCI data and
$+2.1$\,pp on Pen Digits over Euc+$z(T)$, with substantially larger
gains under noise ($+15.1$\,pp and $+12.8$\,pp at $\sigma{=}0.20$).

\textit{Tier~3 (controls and scope).}
We use McNemar tests together with dimension-matched random baselines
($+1$ random dimension against Euc+$z(T)$ and $+15$ random dimensions
against Euc+Heis) to show that the observed improvements arise from
geometric content rather than feature-count inflation.
We also compare the method with ROCKET~\cite{dempster2020}, which
outperforms the proposed descriptor on both datasets and thereby
positions it correctly as a lightweight, interpretable complement to
accuracy-maximising kernel methods rather than a replacement for them.
The evaluation is deliberately restricted to two established online
handwriting benchmarks under clean and Gaussian-noisy conditions, and
the scope of the conclusions is limited accordingly.

%%=========================================================%%
\section{Related Work}
\label{sec:related}
%%=========================================================%%

Research on trajectory and sequence classification spans several
partly overlapping traditions, including time-series classification,
path-signature methods, dynamic time warping, geometric interpolation on
manifolds and Lie groups, and online handwriting recognition.
The present work is most closely connected to all five, but it occupies a
rather specific position within them. It seeks a compact, fixed-length,
order-sensitive descriptor for planar trajectories that remains
interpretable and inexpensive to compute.

In the broader time-series classification literature, strong results have
been achieved through a range of sequence-based methods, including random
convolutional kernels such as ROCKET~\cite{dempster2020},
transformation-based ensembles such as HIVE-COTE~\cite{lines2018,middlehurst2021},
deep convolutional architectures such as
InceptionTime~\cite{ismail_fawaz2020}, and fast shapelet-based
transforms~\cite{cabello2023,sast2024}.
Large-scale comparative studies are given in
\cite{bagnall2017,middlehurst2024}, and a recent survey of deep learning
approaches appears in~\cite{foumani2024}.
These methods are highly effective as general sequence classifiers, but
they are primarily designed for predictive performance rather than for
explicit geometric interpretability.
More importantly for the present study, they do not isolate traversal
order as a first-class geometric quantity.
Our aim is therefore not to compete directly with the strongest general
time-series models, but to address a more specific structural gap: encoding order-sensitivity in planar handwriting trajectories through a low-dimensional, geometrically motivated representation.

A related perspective appears in trajectory analysis outside handwriting,
particularly in surveillance and transport applications.
Work on trajectory clustering~\cite{rodriguez2012} and outlier
detection~\cite{sanroman2019} has shown that motion paths often carry
discriminative information not reducible to static spatial shape alone.
These studies reinforce a point that is also central here. A useful
trajectory descriptor must respect the direction and ordering of motion,
not merely the region of space occupied by the path.
The present paper brings that concern into the domain of online
handwriting and planar pen trajectories, where order-sensitivity has a
particularly direct and interpretable role.

The most immediate mathematical point of contact is the literature on
path signatures.
The truncated path signature provides a systematic representation of a
path through iterated integrals
\cite{chevyrev2022,lyons2026,fermanian2021}, and efficient computation is
available through the \texttt{iisignature}
library~\cite{reizenstein2020}.
Neural controlled differential equations~\cite{kidger2020} extend this
line of work to continuous-time sequence modelling.
For the present paper, the important connection is that the antisymmetric
level-two signature term
$\tfrac{1}{2}(S^{xy}-S^{yx})$ coincides with the terminal signed area
$z(T)$.
In that sense, the global signed area used here is not a new invariant.
The difference lies elsewhere.
The proposed Heisenberg descriptor does not merely retain the endpoint
value of signed area; it builds a compact profile of how the signed area
accumulates along the trajectory and then summarises that profile through
a small number of interpretable statistics.
The contribution is therefore not the identification of the signed area
itself, but its use as the basis for a temporally resolved,
low-dimensional descriptor with measurable empirical benefit.

Dynamic Time Warping provides an important algorithmic baseline from a
different tradition.
DTW combined with nearest-neighbour
classification~\cite{sakoe1978} remains a strong method for online
handwriting because it performs direct sequence matching on the
coordinate stream.
Its strength lies in local shape-sensitive alignment rather than in
fixed-length feature construction.
In our setting, DTW is particularly competitive on short, clean,
fixed-protocol data, but its inference cost grows quadratically in both
dataset size and trajectory length.
This makes it substantially less attractive for large datasets or longer
sequences.
The comparison is therefore useful not because the proposed method is a
drop-in replacement for DTW, but because it clarifies the trade-off.
The Heisenberg descriptor sacrifices some peak clean-data performance in
exchange for compactness, interpretability, and strong robustness in
settings where traversal-order information matters.

The construction of the fuller descriptor also touches the literature on
subdivision and interpolation on manifolds and Lie groups.
A substantial body of work has studied how classical subdivision ideas
can be extended beyond Euclidean vector spaces to Riemannian and
group-valued settings
\cite{wallner2005,grohs2011,park1995,crouch1995,nava2013}.
Subdivision itself belongs to a wider family of constructive methods in
geometric design that generate dense curves and surfaces from sparse
control data.
Related traditions include partial differential equation approaches to
interactive and boundary-driven surface
construction~\cite{ugail1999a,ugail1999b,kubiesa2004}, harmonic and
biharmonic B\'{e}zier patches~\cite{monterde2004,monterde2006}, and
parameterised geometric modelling across visual computing and
engineering~\cite{gonzalez2008,athanasopoulos2009,sheng2010,sheng2011}.
What these approaches share with subdivision is the aim of recovering
controlled dense geometry from a compact description, which is precisely
the operation the present descriptor requires before any statistic is
computed.
Those contributions are primarily concerned with smooth geometric curve
construction, approximation, and interpolation.
The specific noncommutative scheme used here has been analysed as a
geometric object in a companion study, which establishes uniform
convergence of the lifted coordinate and characterises the regularity of
its limit~\cite{ugail2026subdiv}.
The present work draws on that literature at the level of mechanism, but
its objective is different.
Here, subdivision is not used for geometric modelling in its own right.
It is used to produce a stable dense profile from which discriminative
features can be extracted for classification.

Finally, the paper sits within the long-standing literature on online
handwriting recognition~\cite{ghosh2022}.
Earlier feature-engineering approaches demonstrated that compact,
well-chosen descriptors can be highly effective, including SVM-based
systems built on engineered shape features for script-specific character
sets~\cite{shanthi2010,bhattacharya2012} and dimensionality-reduction
approaches for handwritten numeral verification~\cite{zhang2004}.
Related work on handwritten text has also considered retrieval-oriented
problems such as probabilistic word spotting~\cite{toselli2019}.
More recent approaches include multi-task models for joint
classification and trajectory regression~\cite{ott2022}, combinations of
path signatures with pen-tip trajectory features for online character
recognition~\cite{wang2021}, and attention-based sequence models for
handwritten text-line recognition~\cite{kang2022}.
Compared with these methods, the present work deliberately takes a more
restricted but more interpretable route.
It does not attempt end-to-end learning from raw sequences.
Instead, it asks whether a small amount of explicitly order-sensitive
geometry can be added to a conventional Euclidean pipeline in a way that
is simple, fast, and informative.
To our knowledge, the Heisenberg group has not previously been used in
this way as a feature-extraction structure for a planar trajectory
classification.

%%=========================================================%%
\section{Methodology}
\label{sec:method}
%%=========================================================%%

\begin{figure}[t]
\centering
\includegraphics[width=\linewidth]{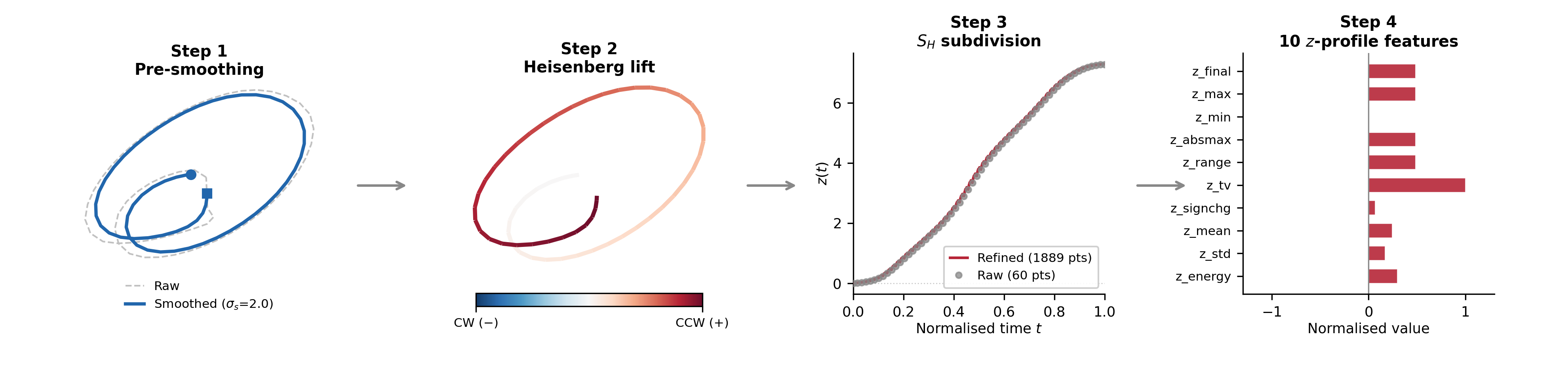}
\caption{The Heisenberg descriptor pipeline applied to one example
  \texttt{o} trajectory ($T{=}60$).
  Step~1: adaptive Gaussian pre-smoothing (raw trajectory in dashed
  grey, smoothed in blue; $\sigma_s{=}2.0$).
  Step~2: Heisenberg lift, each segment coloured by $z(t)$ value
  via a diverging colourmap (red $=$ CW, blue $=$ CCW).
  Step~3: $S_H$ subdivision refines 60 raw control points (grey dots)
  to 1889 values (dark red curve).
  Step~4: ten $z$-profile features, normalised for display;
  \texttt{z\_tv} (total variation) is the largest feature for this
  closed-loop character.
  Total per-curve cost $O(32T)$, approximately $2$\,ms at $T{=}60$.}
\label{fig:pipeline}
\end{figure}

The proposed descriptor is built as a short processing pipeline applied
to each trajectory.
Starting from the raw planar path, the method first reduces the coordinate
noise, then lifts the trajectory into the Heisenberg group so that
signed-area accumulation is represented explicitly, then optionally
refines the lifted path to obtain a denser and more stable profile, and
finally extracts a compact set of summary statistics.
Figure~\ref{fig:pipeline} illustrates these steps for a real
\texttt{o} trajectory of length $T{=}60$.

The motivation for this design is straightforward.
The Euclidean coordinates of a trajectory describe where the path lies in
the plane, but they do not directly encode how oriented area is
accumulated as the trajectory unfolds.
The Heisenberg lift provides a natural way to represent that quantity,
while the subsequent refinement and summarisation steps convert it into a
fixed-length feature vector suitable for standard classifiers.
The descriptor is intentionally layered. In its simplest form it reduces to the terminal signed area alone, whereas in its fuller form it
captures the temporal structure of signed-area accumulation.

\subsection{Adaptive Gaussian Pre-Smoothing}

The first step is a light smoothing of the coordinate sequence.
This is necessary because the lifted coordinate defined in
Eq.~\eqref{eq:lift} is a cumulative sum of products of adjacent
coordinates and therefore accumulates noise as the path progresses.
If additive Gaussian noise
$\varepsilon \sim \mathcal{N}(0,\sigma^2)$
is applied independently to the coordinate channels, the variance of the
terminal lifted value grows approximately as,
\begin{equation}
  \mathrm{Var}(z_T) \;\approx\; T \cdot \sigma^2
    \cdot \mathbb{E}[x^2 + y^2],
  \label{eq:variance}
\end{equation}
so longer trajectories are increasingly affected.
Because the subdivision scheme used later preserves existing nodes and
inserts new ones without denoising them. Noise suppression must therefore be performed before the lift rather than afterwards.

We therefore apply a one-dimensional Gaussian filter independently to
the two coordinate channels, using an adaptive bandwidth,
\begin{equation}
  \sigma_s = \max\!\bigl(0.3,\; 2.0 \cdot T / 60\bigr),
  \label{eq:adaptive}
\end{equation}
which scales with trajectory length.
The purpose of this choice is to keep smoothing mild on short
trajectories while allowing stronger regularisation when more densely
sampled paths are available.
At $T{=}60$ this yields $\sigma_s{=}2.0$, whereas at $T{=}8$ it yields
$\sigma_s{=}0.30$.
Using a fixed larger value on very short trajectories suppresses not
only noise, but also the discriminative geometry of the stroke; for
example, a fixed $\sigma_s{=}2.0$ at $T{=}8$ reduces classification
accuracy from $0.952$ to $0.840$.

\subsection{Heisenberg Lift}

After smoothing, the planar trajectory is lifted into the Heisenberg
group $\Heis = (\R^3,\cdot)$, whose multiplication law is,
\begin{equation}
  (x,y,z)\cdot(x',y',z') =
  \bigl(x+x',\; y+y',\; z+z'+\tfrac{1}{2}(xy'-yx')\bigr).
  \label{eq:grouplaw}
\end{equation}
The cross-term $\tfrac{1}{2}(xy'-yx')$ is the signed area of the
parallelogram spanned by $(x,y)$ and $(x',y')$.
Its antisymmetry is precisely what makes it suitable for encoding
traversal direction, reversing the order of the inputs reverses the sign
of the area contribution.
This group is two-step nilpotent and admits an explicit polynomial
exponential map, which makes the resulting computations simple and
closed-form.

For a discrete trajectory $(x_t,y_t)$, the lift is defined by the
recursion,
\begin{equation}
  z_0 = 0,\quad z_{t+1} = z_t
    + \tfrac{1}{2}(x_t y_{t+1} - y_t x_{t+1}),
  \label{eq:lift}
\end{equation}
so that the third coordinate records the cumulative signed-area
accumulation along the path.
The terminal value $z(T)$ already provides a useful scalar descriptor.
For a closed counter-clockwise loop of radius $r$, it is approximately
$+\pi r^2$; for a clockwise loop it is approximately $-\pi r^2$.
For open curves, it measures the net oriented area swept out by the
trajectory.
In all cases, it captures information that is invisible to standard
Euclidean shape statistics.

A useful structural property of the lift is that it respects
concatenation.
If a trajectory is split into two consecutive segments, the lift of the
full path equals the group product of the lifts of the two segments.
This compatibility is important because it allows subsequent refinement
to remain faithful to the signed-area interpretation.

\subsection{$S_H$ Noncommutative Subdivision}
\label{sec:subdivision}

The terminal signed area alone is often already informative, but it does
not describe how the area accumulated over time.
To obtain a denser and more stable temporal profile, the lifted polygon
is refined by a Heisenberg-group four-point subdivision scheme, denoted
by $S_H$.
The scheme is interpolatory, meaning that all existing control points are
preserved and new points are inserted between them.
Its horizontal coordinates are generated by the classical four-point
operator $L_4$~\cite{dyn1987,dyn1992,dyn2002}, while the lifted
coordinate is corrected to remain consistent with the noncommutative
group law,
\begin{equation}
  Z_{2i+1} = (L_4\, z^k)_{2i+1} + R^k_i, \qquad
  R^k_i = \tfrac{1}{2}(x^k_i\, b^k_i - y^k_i\, a^k_i),
  \label{eq:insertion}
\end{equation}
where $a^k_i$ and $b^k_i$ are the horizontal displacements to the new
node and $R^k_i$ is the noncommutative correction term.

The correction is uniformly bounded by $|R^k_i| \leq A \cdot 2^{-k}$,
which is sufficient to guarantee that the refined lifted coordinate
converges uniformly to a continuous limit and that all original control
points are preserved exactly~\cite{ugail2026subdiv,dyn2002}.
It does not, however, deliver a continuously differentiable limit.
The scaled correction satisfies
$2^k R^k_i \rightarrow \tfrac{1}{4}(X(t)Y'(t) - Y(t)X'(t))$, so the
forcing that drives the first differences of the lifted coordinate does
not decay with the refinement level.
The horizontal coordinates therefore inherit the classical $C^1$
four-point limit exactly, while the lifted coordinate converges to a
limit in the Zygmund class with modulus of continuity of order
$h \log(1/h)$, which under an explicit non-cancellation condition fails
to be continuously differentiable~\cite{ugail2026subdiv}.
This distinction does not affect the descriptor constructed here.
What the refinement is required to supply is a dense and uniformly
convergent estimate of the signed-area profile on a preserved node set,
and the statistics of Table~\ref{tab:features} are integral and
variational summaries of that profile rather than pointwise derivative
quantities.
The practical point is therefore not smoothness for its own sake, but the
fact that the refined curve provides a stable estimate of the signed-area
profile at a much higher sampling density.

The effect of this refinement depends strongly on trajectory length.
After $n{=}5$ refinement steps, the number of samples increases by a
factor of $32$, so a trajectory with $T$ control points yields a profile
of $32(T-1)+1$ values.
For example, a $T{=}60$ path produces 1889 profile values.
At such lengths, the unrefined profile is already fairly dense, so the
benefit of subdivision on clean data is limited.
At $T{=}8$, however, the raw profile is sparse, and the refined version
provides a more stable basis for estimating variation and other temporal
statistics.
Its practical contribution is therefore most visible under noise and in
short or loop-heavy trajectories, rather than in already well-sampled
clean cases.

\subsection{Feature Extraction}

The final step converts the lifted trajectory into a fixed-length
descriptor.
From the refined Heisenberg curve we extract fifteen summary statistics,
listed in Table~\ref{tab:features}.
The first ten describe the signed-area profile and are explicitly
order-sensitive.
The remaining five capture complementary horizontal geometry.
This separation reflects the intended role of the descriptor. It is not
meant to discard ordinary shape information, but to add order-sensitive
structure to it.

\begin{table}[t]
\caption{Heisenberg feature vector (15 dimensions).
  Features~1--10 are order-sensitive; features~11--15 describe
  horizontal geometry.}
\label{tab:features}
\centering\small
\begin{tabular}{clp{5.0cm}}
\toprule
\# & Name & Description \\
\midrule
1  & \texttt{z\_final}   & Terminal signed area $z(T)$ \\
2  & \texttt{z\_max}     & Maximum accumulated area \\
3  & \texttt{z\_min}     & Minimum accumulated area \\
4  & \texttt{z\_absmax}  & Maximum absolute excursion \\
5  & \texttt{z\_range}   & Total excursion range \\
6  & \texttt{z\_tv}      & Total variation of $z$ \\
7  & \texttt{z\_signchg} & Number of sign changes in $z$ \\
8  & \texttt{z\_mean}    & Mean of $z$ along path \\
9  & \texttt{z\_std}     & Standard deviation of $z$ \\
10 & \texttt{z\_energy}  & Energy $\sum_t z_t^2$ \\
11 & \texttt{h\_length}  & Horizontal arc length \\
12 & \texttt{h\_disp}    & Endpoint displacement \\
13 & \texttt{h\_curv}    & Mean curvature proxy \\
14 & \texttt{z\_slope}   & $z(T)$ per unit arc length \\
15 & \texttt{z\_skew}    & Skewness of the $z$-profile \\
\bottomrule
\end{tabular}
\end{table}

Feature~1, the terminal signed area $z(T)$, is the simplest and often
most informative quantity.
Features~2--10 describe how that area is accumulated: whether the
profile is monotone or oscillatory, whether it undergoes reversals,
whether its mass is concentrated early or late, and how large its total
variation is.
These additional statistics become useful when the terminal scalar alone
does not capture enough of the within-class variation.
The hard binary pair \texttt{o} versus \texttt{y} provides a clear
illustration.
For that pair, $z(T)$ alone already achieves perfect separation,
\texttt{o} generates a large, mostly monotone area-accumulation profile,
whereas \texttt{y} remains close to zero because its path is open and
doubles back.
In such a case a single scalar is sufficient.
In noisier or more diverse multi-class settings, however, the fuller
profile contributes additional discriminative information.

The final combined descriptor is obtained by concatenating the
15-dimensional Heisenberg vector with a 25-dimensional Euclidean
baseline consisting of arc length, endpoint displacement, mean
curvature, and 22 discrete Fourier amplitudes.
This yields the 40-dimensional Euc+Heis representation used in the main
experiments.
A lighter alternative is Euc+$z(T)$, in which only the terminal signed
area is appended.
The experiments below are designed to show when the minimal scalar is
already enough and when the fuller profile provides a worthwhile gain.

%%=========================================================%%
\section{Experimental Results}
\label{sec:experiments}
%%=========================================================%%

This section evaluates the proposed descriptor from three complementary
angles.
We first describe the experimental protocol, baselines, and datasets.
We then use a controlled synthetic task to verify that the method
responds to the order-sensitive signal it is designed to capture.
Finally, we turn to two real handwriting benchmarks and examine clean
classification performance, a hard same-shape confusion, and robustness
under additive noise.

\subsection{Experimental Protocol, Baselines, and Data}

All experiments use five-fold stratified cross-validation with a fixed
random seed of 42.
Two classifiers are evaluated throughout. A Random Forest with 150 trees
and an RBF-kernel SVM with $C{=}10$, each preceded by
\texttt{StandardScaler}.
The intention is not to optimise classifier design, but to assess the
behaviour of the proposed features under two standard and widely used
classification models.

To assess whether observed improvements are statistically meaningful, we
apply McNemar's corrected test~\cite{mcnemar1947,demsar2006} with
continuity correction and one degree of freedom to pooled
cross-validation predictions.
Predictions from the five non-overlapping test folds are concatenated
into a single vector, and the corrected test is applied to the resulting
contingency table of correct and incorrect predictions.
This allows significance to be assessed at the level of disagreement
patterns between methods rather than by comparing mean accuracies alone.

\begin{figure}[t]
\centering
\includegraphics[width=\linewidth]{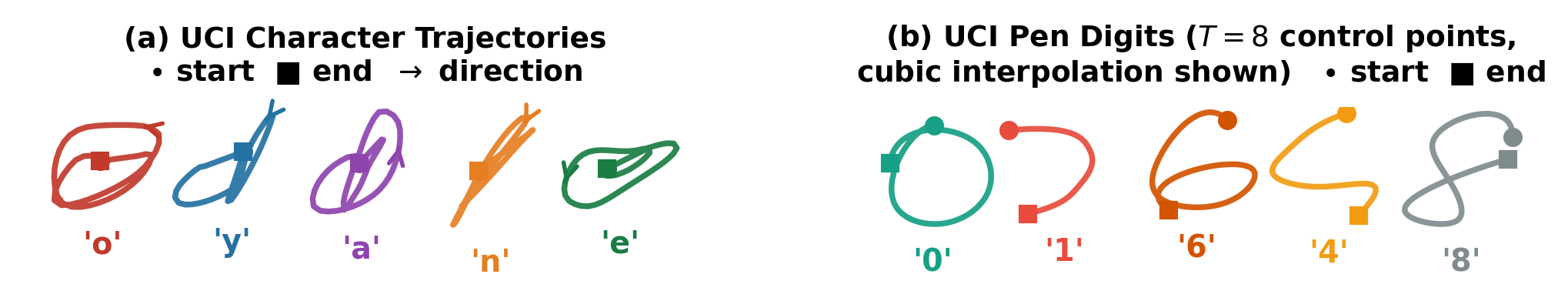}
\caption{Representative trajectory examples.
  Panel~(a): five characters from UCI Character Trajectories ($T{=}60$):
  \texttt{o}, \texttt{y}, \texttt{a}, \texttt{n}, \texttt{e}.
  Filled circles mark starts, squares mark ends, arrows indicate direction.
  Note the contrasting loop structures of \texttt{o} (closed, positive $z(T)$)
  and \texttt{y} (open, near-zero $z(T)$).
  Panel~(b): five digits from UCI Pen Digits ($T{=}8$ control points,
  cubic interpolation shown): \texttt{0}, \texttt{1}, \texttt{6},
  \texttt{4}, \texttt{8}.
  Loop-containing digits (0, 6, 8) generate larger signed-area signals
  than open strokes (1, 4).}
\label{fig:datasets}
\end{figure}

The two real datasets used in the study differ in both trajectory length
and recording protocol.
UCI Character Trajectories~\cite{williams2006} contains 2858 online
handwriting trajectories from 20 character classes, recorded originally
as pen-tip velocities.
These are integrated to Cartesian positions, resampled to a common
length of $T{=}60$, and normalised to zero mean and unit scale on a
per-curve basis.
This dataset is particularly appropriate for the present study because
many of its classes share similar spatial outlines and differ mainly in
stroke direction or sequencing, which is precisely the information that
Euclidean descriptors fail to represent.
UCI Pen-Based Digit Recognition~\cite{alpaydin1998} contains 10,992
samples from 10 digit classes written by 44 writers.
Each sample consists of only $T{=}8$ control points, making it
structurally different from the character dataset and especially useful
for evaluating the descriptor in the short-trajectory regime.
Representative examples from both datasets are shown in
Fig.~\ref{fig:datasets}.

Writer independence can be verified directly only for UCI Pen Digits,
which is distributed with a predefined writer-independent split of 7494
training and 3498 test samples.
We compare our five-fold cross-validation results against this split and
find differences of at most $3.9$ percentage points, which is
consistent with the smaller test set and expected sampling variation.
UCI Character Trajectories does not provide writer identifiers in its
released form, so five-fold stratified cross-validation is used in line
with common practice for that dataset~\cite{bagnall2017,middlehurst2024}.

The evaluation includes several baselines chosen to test different
aspects of the proposed method.
The first is a conventional Euclidean descriptor based on path
statistics and Fourier amplitudes.
The second consists of truncated path signatures at levels 2 and 3.
The third is DTW with one-nearest-neighbour classification, which
provides a strong sequence-matching baseline.
We also include ROCKET~\cite{dempster2020}, using 1000 random
convolutional kernels and a Ridge classifier, as a strong modern
reference point.
This is not intended as a directly comparable method in terms of
interpretability or feature dimension, but as an upper-bound indicator of
what a high-capacity time-series classifier can achieve.
The evaluation further includes two random augmentation controls: Euc+Rand, which matches Euc+$z(T)$ in dimension, and Euc+Rand$_{15}$, which matches Euc+Heis in dimension.
These null baselines are important because they allow improvements to be
attributed to geometric information rather than to an increase in
feature count.
Table~\ref{tab:methods} summarises all feature methods used in the
experiments.

\begin{table}[t]
\caption{Feature methods evaluated. Ablation variants are italicised.
  $\dagger$ indicates a dimension-controlled null baseline:
  Euc+Rand matches Euc+$z(T)$ in dimension (26);
  Euc+Rand$_{15}$ matches Euc+Heis in dimension (40).}
\label{tab:methods}
\centering\small
\begin{tabular}{llc}
\toprule
Label & Description & Dim \\
\midrule
Euc          & Euclidean: path stats + DFT amplitudes & 25 \\
Sig~L2       & Truncated path signature, level~2~\cite{chevyrev2022} & 6 \\
Sig~L3       & Truncated path signature, level~3 & 14 \\
DTW          & DTW + 1-NN~\cite{sakoe1978} & --- \\
\midrule
\textit{z(T) only}   & Terminal signed area alone & 1 \\
\textit{Euc+z(T)}    & Euc concatenated with $z(T)$ & 26 \\
\textit{Heis-noSH}   & Heis lift without subdivision & 15 \\
Heis                 & Heis with adaptive smoothing and $S_H$ & 15 \\
Euc+Rand$^\dagger$       & Euc + 1 random Gaussian projection & 26 \\
Euc+Rand$_{15}$$^\dagger$ & Euc + 15 random Gaussian projections & 40 \\
Euc+Heis             & Euc + Heis (proposed combined) & 40 \\
\bottomrule
\end{tabular}
\end{table}

Because DTW has inference cost of order $O(N^2T^2)$, its use depends
strongly on dataset size and trajectory length.
It is therefore evaluated on the full pen-digit dataset, where
$T{=}8$ makes computation tractable, but on UCI Character Trajectories,
where $T{=}60$ and $N{=}2858$, it is reported from a 500-curve
stratified subsample and is marked accordingly.
The complete implementation is provided in a Google Colab notebook in
the supplementary material.

\subsection{Controlled Orientation-Sensitivity Experiment}
\label{sec:exp:synthetic}

Before turning to real handwriting data, we first verify that the
pipeline actually captures the specific signal it was designed to
extract.
To do so, we construct a binary synthetic task in which traversal
direction is the only discriminating information.
The dataset consists of 200 clockwise and 200 counter-clockwise loops,
each sampled at $T{=}64$ points, with random radius, centre, and start
angle.
By construction, all Euclidean shape statistics are identical between
the two classes.

This experiment is intended as a mechanistic validation rather than as
evidence of real-world superiority.
On clean data, shown in Table~\ref{tab:synth_clean}, all explicitly
order-sensitive methods achieve perfect or near-perfect accuracy, whereas
Euclidean features and their random augmentation remain far below that
level.
The result confirms that once the traversal direction is the only useful
signal, the signed-area representation captures it exactly.
The noise experiment in Table~\ref{tab:synth_noise} shows the same
pattern more strongly.
Euclidean performance collapses towards chance almost immediately,
whereas the Heisenberg variants with adaptive pre-smoothing remain
perfect through the full noise range tested.
This provides an empirical check on the motivation for the adaptive
smoothing rule in Eq.~\eqref{eq:adaptive}: the lifted coordinate is
stable under noise only when smoothing is applied before the lift.

\begin{table}[t]
\caption{Controlled orientation task, clean data (RF/SVM accuracy,
  5-fold CV).}
\label{tab:synth_clean}
\centering\small
\begin{tabular}{lccc}
\toprule
Method & RF & SVM & F1 \\
\midrule
Euc                   & $0.838\pm0.030$ & $0.845\pm0.026$ & 0.837 \\
Sig~L2                & $1.000\pm0.000$ & $0.998\pm0.005$ & 1.000 \\
\textit{z(T) only}    & $1.000\pm0.000$ & $1.000\pm0.000$ & 1.000 \\
\textit{Euc+z(T)}     & $0.998\pm0.005$ & $0.998\pm0.005$ & 0.998 \\
Heis (adaptive)       & $1.000\pm0.000$ & $1.000\pm0.000$ & 1.000 \\
Euc+Rand              & $0.818\pm0.041$ & $0.845\pm0.026$ & 0.817 \\
\textbf{Euc+Heis}     & $\mathbf{1.000\pm0.000}$ & $\mathbf{1.000\pm0.000}$ & 1.000 \\
\bottomrule
\end{tabular}
\end{table}

\begin{table}[t]
\caption{Controlled orientation task, noise robustness (RF accuracy).}
\label{tab:synth_noise}
\centering\small
\begin{tabular}{lcccccc}
\toprule
Method & $\sigma{=}0$ & $0.05$ & $0.10$ & $0.15$ & $0.20$ & $0.30$ \\
\midrule
Euclidean       & 0.838 & 0.522 & 0.512 & 0.510 & 0.503 & 0.492 \\
Sig~L2          & 1.000 & 1.000 & 1.000 & 1.000 & 1.000 & 0.993 \\
Heis (no smooth)& 1.000 & 1.000 & 1.000 & 1.000 & 1.000 & 0.993 \\
Heis (adaptive) & 1.000 & 1.000 & 1.000 & 1.000 & 1.000 & \textbf{1.000} \\
\textbf{Euc+Heis} & 1.000 & 1.000 & 1.000 & 1.000 & 1.000 & \textbf{1.000} \\
\bottomrule
\end{tabular}
\end{table}

\subsection{UCI Character Trajectories}
\label{sec:exp:uci}

We now turn to the first real benchmark, UCI Character Trajectories.
This dataset is the more demanding of the two in terms of length and
within-class variation, and it is also the setting in which the argument
for order-sensitive information is easiest to motivate.

Table~\ref{tab:uci_20class} reports clean-data results on the full
20-class task.
The central result is the performance of Euc+$z(T)$.
Appending the terminal signed area to the Euclidean baseline increases
Random Forest accuracy from $0.833$ to $0.889$ and SVM accuracy from
$0.868$ to $0.909$.
This improvement is statistically significant and, crucially, it is not
replicated by the dimension-matched random baseline, which remains
essentially identical to Euclidean performance.
The result shows that a single additional scalar can recover useful
information missing from the Euclidean descriptor.
At the same time, $z(T)$ alone performs poorly in isolation, which makes
clear that the signed-area feature should be understood as a complement
to shape information rather than as a standalone replacement for it.

The full Heisenberg descriptor performs slightly better still, reaching
$0.901$ with Random Forest and $0.925$ with SVM.
That additional gain over Euc+$z(T)$ is statistically significant, but
it is much smaller than the gain of Euc+$z(T)$ over Euclidean alone.
This distinction is important.
It indicates that the principal practical contribution of the method lies
in the scalar augmentation, while the richer temporal profile should be
viewed as an extension that becomes worthwhile under specific operating
conditions.

\begin{table}[t]
\caption{UCI Character Trajectories, 20-class, clean data.
  Bold: proposed core method Euc+$z(T)$ (26-dim).
  $\star$: dimension-controlled null.
  $\dagger$: DTW from 500-curve subsample (not directly comparable).
  $\ddagger$: ROCKET (Ridge, 1k kernels), reference upper bound.}
\label{tab:uci_20class}
\centering\small
\begin{tabular}{lcccc}
\toprule
Method & Dim & RF & SVM & F1 \\
\midrule
\multicolumn{5}{l}{\textit{Reference baselines}} \\
Euc               & 25 & $0.833\pm0.014$ & $0.868\pm0.016$ & 0.832 \\
Sig~L2            &  6 & $0.659\pm0.009$ & $0.701\pm0.015$ & 0.658 \\
Sig~L3            & 14 & $0.842\pm0.010$ & $0.856\pm0.016$ & 0.840 \\
DTW$^\dagger$     &--- & \multicolumn{2}{c}{$0.964\pm0.008$} & ---   \\
ROCKET$^\ddagger$ & 2k & \multicolumn{2}{c}{$0.975\pm0.006$} & ---   \\
\midrule
\multicolumn{5}{l}{\textit{Core contribution: Euc augmented with z(T)}} \\
\textit{z(T) only}    &  1 & $0.255\pm0.013$ & $0.335\pm0.015$ & ---  \\
Euc+Rand$^\star$      & 26 & $0.833\pm0.009$ & $0.861\pm0.014$ & 0.832 \\
\textbf{Euc+$z(T)$}  & 26 & $\mathbf{0.889\pm0.016}$
                           & $\mathbf{0.909\pm0.011}$ & \textbf{0.888} \\
\midrule
\multicolumn{5}{l}{\textit{Extension: full Heisenberg profile}} \\
Heis-noSH     & 15 & $0.867\pm0.013$ & $0.887\pm0.011$ & 0.867 \\
Heis (adap.)  & 15 & $0.858\pm0.017$ & $0.878\pm0.012$ & 0.858 \\
Euc+Heis      & 40 & $0.901\pm0.008$ & $0.925\pm0.008$ & 0.901 \\
\bottomrule
\end{tabular}
\end{table}

To make this point more concrete, we next examine the hardest same-shape
confusion identified by screening all character pairs for cases in which
Euclidean accuracy falls below $0.93$.
The most difficult pair is \texttt{o} versus \texttt{y}, shown in
Table~\ref{tab:binary}.
This pair is geometrically revealing.
The character \texttt{o} is a closed loop and therefore generates
substantial net signed area, whereas \texttt{y} is an open stroke whose
net area remains close to zero.
Euclidean features cannot represent that distinction and plateau at
$0.964$.
A random augmentation does not improve the result.
By contrast, $z(T)$ alone reaches perfect accuracy.
Once the scalar is present, the richer temporal statistics add nothing
further on this specific pair.
This is exactly the behaviour one would hope to see: when the
discriminating signal collapses to a single order-sensitive quantity, the
minimal version of the method is sufficient.

\begin{table}[t]
\caption{Hard binary task: \texttt{o} vs \texttt{y}, UCI Character
  Trajectories ($n{=}141$ per class, RF accuracy, 5-fold CV).
  $z(T)$ alone (1-dim) achieves $1.000$ while 25-dim Euclidean reaches
  only $0.964$.}
\label{tab:binary}
\centering\small
\begin{tabular}{lcc}
\toprule
Method & RF Acc & RF $\pm$ \\
\midrule
Euc                & 0.964 & 0.032 \\
Euc+Rand           & 0.960 & 0.027 \\
\textit{z(T) only} & 1.000 & 0.000 \\
Sig~L3             & 0.852 & 0.041 \\
\textit{Euc+z(T)}  & 1.000 & 0.000 \\
\textit{Heis-noSH} & 1.000 & 0.000 \\
\textbf{Euc+Heis}  & \textbf{1.000} & 0.000 \\
\bottomrule
\end{tabular}
\end{table}

\begin{figure}[t]
\centering
\includegraphics[width=\linewidth]{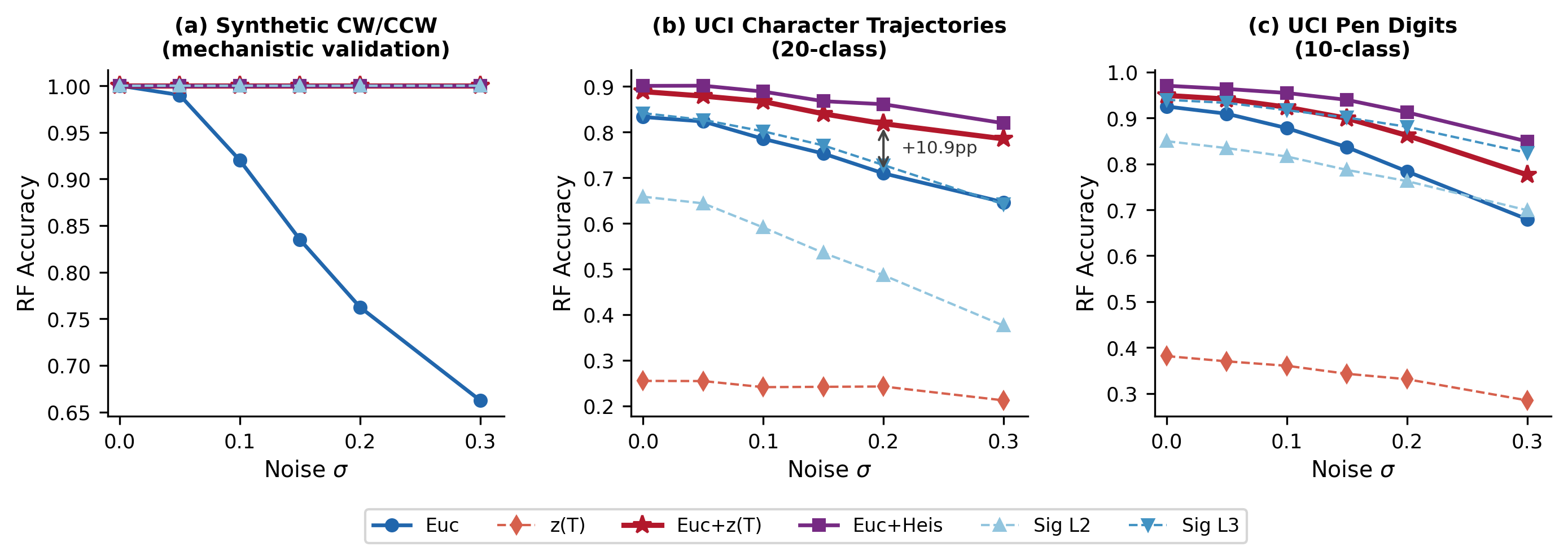}
\caption{Noise robustness (RF accuracy under additive Gaussian noise
  $\sigma \in \{0,0.05,0.10,0.15,0.20,0.30\}$).
  Panel~(a): synthetic CW/CCW task (mechanistic validation).
  Panel~(b): UCI Character Trajectories (20-class); Euc+$z(T)$ leads
  Euclidean by $+10.9$\,pp at $\sigma{=}0.20$ (annotated arrow);
  Euc+Heis extends this to $+15.1$\,pp.
  Panel~(c): UCI Pen Digits (10-class); Euc+$z(T)$ gains $+7.8$\,pp
  and Euc+Heis gains $+12.8$\,pp at $\sigma{=}0.20$.
  Shared legend below all panels.}
\label{fig:noise}
\end{figure}

The final experiment on this dataset examines robustness under additive
Gaussian coordinate noise.
Table~\ref{tab:uci_noise} shows that the advantage of the proposed
descriptor grows as the trajectories become more corrupted.
At $\sigma{=}0.20$, Euc+$z(T)$ leads Euclidean by $10.9$ percentage
points, and the full Euc+Heis descriptor widens this margin to
$15.1$ percentage points.
This is one of the settings in which the fuller temporal profile becomes
most useful.
The benefit of adaptive pre-smoothing is also directly visible. The
difference between Heisenberg features computed without pre-smoothing
and those computed with it confirms the practical importance of the
variance argument in Eq.~\eqref{eq:variance}.
Figure~\ref{fig:noise} summarises the noise behaviour across all three
benchmark settings.

\begin{table}[t]
\caption{UCI Character Trajectories noise robustness (RF accuracy,
  20-class). Noise $\sigma$ is relative to unit-std normalised
  trajectories.}
\label{tab:uci_noise}
\centering\small
\begin{tabular}{lcccccc}
\toprule
Method & $0$ & $0.05$ & $0.10$ & $0.15$ & $0.20$ & $0.30$ \\
\midrule
Euclidean           & 0.833 & 0.823 & 0.785 & 0.753 & 0.710 & 0.646 \\
Sig~L2              & 0.659 & 0.644 & 0.591 & 0.535 & 0.486 & 0.376 \\
Sig~L3              & 0.841 & 0.827 & 0.802 & 0.771 & 0.728 & 0.642 \\
\textbf{Euc+$z(T)$} & \textbf{0.889} & \textbf{0.879}
  & \textbf{0.867} & \textbf{0.840} & \textbf{0.818} & \textbf{0.785} \\
Euc+Heis & 0.901 & 0.902 & 0.889 & 0.868 & 0.861 & 0.819 \\
\bottomrule
\end{tabular}
\end{table}

\subsection{UCI Pen-Based Digit Recognition}
\label{sec:exp:pendigits}

The second real benchmark, UCI Pen Digits, provides a rather different
test case.
Trajectories are much shorter, the recording protocol is cleaner, and
several classes have a strong loop structure.
This makes the dataset useful for testing whether the fuller descriptor
becomes more valuable when only a sparse control polygon is available.

Table~\ref{tab:pendigits} reports the clean-data results.
The overall pattern is consistent with that seen on UCI Character
Trajectories, but the quantitative balance shifts.
Euc+$z(T)$ again improves on the Euclidean baseline, reaching $0.949$
with Random Forest and $0.963$ with SVM.
The gain is highly significant and is not reproduced by the random
baseline, which remains slightly below Euclidean performance.
The full Euc+Heis descriptor improves further to $0.970$, exceeding
Euc+$z(T)$ by $2.1$ percentage points.
This larger gap is consistent with the structure of the dataset.
Digits such as 0, 6, 8, and 9 contain strong loop geometry, and the
richer signed-area profile is therefore more informative here than on the
longer, more variable character dataset.
DTW remains the best-performing method on this benchmark, which is not
surprising given the short, clean, fixed-protocol nature of the data.

\begin{table}[t]
\caption{UCI Pen Digits, 10-class, clean data.
  $\ddagger$: ROCKET (Ridge, 1k kernels), reference upper bound.}
\label{tab:pendigits}
\centering\small
\begin{tabular}{lcccc}
\toprule
Method & Dim & RF & SVM & F1 \\
\midrule
\multicolumn{5}{l}{\textit{Reference baselines}} \\
Euc               & 25 & $0.925\pm0.008$ & $0.942\pm0.004$ & 0.925 \\
Sig~L2            &  6 & $0.849\pm0.007$ & $0.838\pm0.007$ & 0.849 \\
Sig~L3            & 14 & $0.940\pm0.005$ & $0.939\pm0.006$ & 0.940 \\
DTW               &---& \multicolumn{2}{c}{$0.994\pm0.001$}  & ---   \\
ROCKET$^\ddagger$ & 2k& \multicolumn{2}{c}{$0.995\pm0.001$}  & ---   \\
\midrule
\multicolumn{5}{l}{\textit{Core contribution: Euc augmented with z(T)}} \\
\textit{z(T) only}   &  1 & $0.381\pm0.008$ & $0.489\pm0.006$ & ---  \\
Euc+Rand$^\star$     & 26 & $0.921\pm0.006$ & $0.934\pm0.006$ & 0.921 \\
\textbf{Euc+$z(T)$} & 26 & $\mathbf{0.949\pm0.003}$
                          & $\mathbf{0.963\pm0.005}$ & \textbf{0.949} \\
\midrule
\multicolumn{5}{l}{\textit{Extension: full Heisenberg profile}} \\
Heis-noSH     & 15 & $0.945\pm0.004$ & $0.958\pm0.005$ & 0.945 \\
Heis (adap.)  & 15 & $0.949\pm0.003$ & $0.960\pm0.004$ & 0.949 \\
Euc+Heis      & 40 & $0.970\pm0.002$ & $0.984\pm0.002$ & 0.970 \\
\bottomrule
\end{tabular}
\end{table}

Noise robustness on Pen Digits is reported in
Table~\ref{tab:pendigits_noise} and in panel~(c) of
Fig.~\ref{fig:noise}.
Here too the proposed descriptor becomes more advantageous as the noise
level increases.
At $\sigma{=}0.20$, Euc+$z(T)$ improves on the Euclidean baseline by
$7.8$ percentage points, while Euc+Heis improves by $12.8$.
The margin between Euc+Heis and Euc+$z(T)$ is also larger than on the
character dataset, again supporting the view that the fuller
signed-area profile is particularly useful for short, loop-structured
trajectories in which the raw temporal sampling is sparse.

\begin{table}[t]
\caption{UCI Pen Digits, noise robustness (RF accuracy, 10-class).}
\label{tab:pendigits_noise}
\centering\small
\begin{tabular}{lcccccc}
\toprule
Method & $0$ & $0.05$ & $0.10$ & $0.15$ & $0.20$ & $0.30$ \\
\midrule
Euclidean           & 0.925 & 0.909 & 0.878 & 0.836 & 0.784 & 0.679 \\
Sig~L2              & 0.849 & 0.834 & 0.816 & 0.787 & 0.763 & 0.699 \\
Sig~L3              & 0.940 & 0.933 & 0.917 & 0.901 & 0.881 & 0.824 \\
\textbf{Euc+$z(T)$} & \textbf{0.949} & \textbf{0.941} & \textbf{0.923}
  & \textbf{0.899} & \textbf{0.862} & \textbf{0.776} \\
Euc+Heis            & 0.970 & 0.963 & 0.955 & 0.940 & 0.912 & 0.849 \\
\bottomrule
\end{tabular}
\end{table}

%%=========================================================%%
\section{Discussion}
\label{sec:discussion}
%%=========================================================%%

\begin{figure}[t]
\centering
\includegraphics[width=\linewidth]{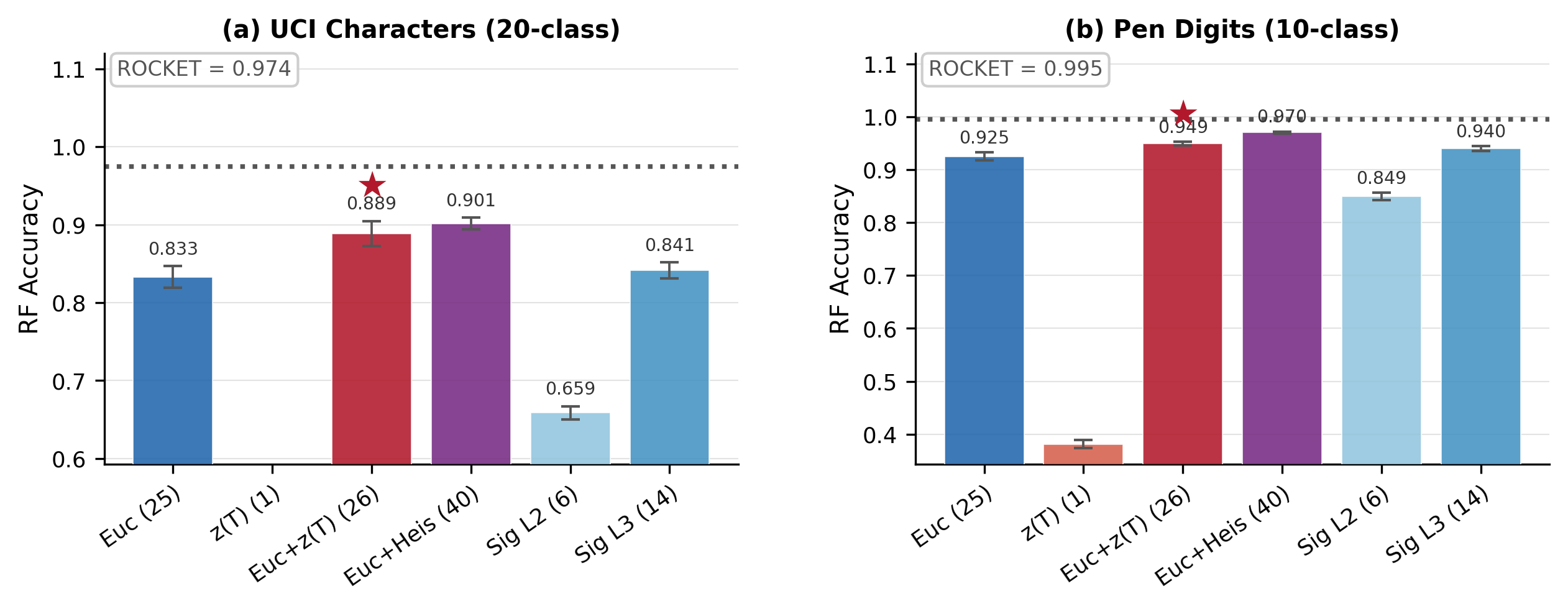}
\caption{Main clean-data results (RF accuracy, 5-fold CV,
  error bars $= \pm1$ std).
  Panel~(a): UCI Character Trajectories (20-class).
  Panel~(b): Pen Digits (10-class).
  The starred bar ($\star$, dark red) is the proposed Euc+$z(T)$
  augmentation (26-dim, one extra parameter-free scalar).
  The dotted reference line marks the ROCKET upper bound
  ($0.974$ on UCI; $0.995$ on Pen Digits).
  Euc+Rand (same 26-dim null baseline) matches Euclidean, confirming
  that the gain is geometric rather than dimensional.}
\label{fig:ablation}
\end{figure}

Table~\ref{tab:ablation} consolidates the main ablation results across
both real benchmarks, Fig.~\ref{fig:ablation} summarises the clean-data
performance, and Table~\ref{tab:mcnemar} reports the corresponding
McNemar tests.
Taken together, these results support three main conclusions.
First, the terminal signed area plays the central practical role in the
method.
Second, the fuller Heisenberg profile adds value primarily in noisier
and more structurally demanding settings.
Third, the observed gains are attributable to geometric information
rather than to increased feature dimensionality.

\begin{table}[t]
\caption{Component ablation: RF accuracy and difference relative to
  Euclidean baseline ($\Delta$).}
\label{tab:ablation}
\centering\small
\begin{tabular}{lcccc}
\toprule
 & \multicolumn{2}{c}{UCI Characters}
 & \multicolumn{2}{c}{Pen Digits} \\
\cmidrule(lr){2-3}\cmidrule(lr){4-5}
Method & RF & $\Delta$ & RF & $\Delta$ \\
\midrule
Euc (baseline)           & 0.8331 & ---      & 0.9250 & ---    \\
\midrule
\multicolumn{5}{l}{\textit{(A) Is z(T) alone sufficient?}}\\
\quad \textit{z(T) only} & 0.255  & $-0.578$ & 0.381  & $-0.544$ \\
\quad \textit{Euc+z(T)}  & 0.889  & $+0.056$ & 0.949  & $+0.024$ \\
\quad Heis (15-dim alone) & 0.858  & $+0.025$ & 0.949  & $+0.024$ \\
\quad Euc+Heis            & 0.901  & $+0.068$ & 0.970  & $+0.045$ \\
\midrule
\multicolumn{5}{l}{\textit{(B) Does $S_H$ subdivision add value?}}\\
\quad \textit{Heis-noSH} & 0.867  & ---      & 0.945  & ---    \\
\quad Heis+SH (adaptive) & 0.858  & $-0.009$ & 0.949  & $+0.024$ \\
\midrule
\multicolumn{5}{l}{\textit{(C) Dimensions or geometric information?}}\\
\quad Euc+Rand (26-dim)    & 0.833  & $+0.002$ & 0.921  & $-0.004$ \\
\quad Euc+Rand$_{15}$ (40-dim) & 0.835  & $+0.003$ & 0.924  & $-0.001$ \\
\quad Euc+Heis (40-dim)    & 0.901  & $+0.068$ & 0.970  & $+0.045$ \\
\bottomrule
\end{tabular}
\end{table}

The ablation results clarify the internal structure of the method.
The scalar signed area is not, by itself, a general-purpose classifier:
without Euclidean shape context, $z(T)$ alone reaches only $0.255$ on
UCI and $0.381$ on Pen Digits.
Its role is therefore complementary rather than substitutive.
Once combined with the Euclidean baseline, however, Euc+$z(T)$ becomes
the main source of the observed improvement, raising accuracy to
$0.889$ on UCI and $0.949$ on Pen Digits.
This simple one-feature augmentation recovers most of the gap between
Euclidean features and the full Heisenberg descriptor.
The remaining gap between Euc+$z(T)$ and Euc+Heis is modest on clean
data, namely $+1.3$\,pp on UCI and $+2.1$\,pp on Pen Digits, which
suggests that the richer temporal profile is useful, but not the primary
driver in the clean benchmark setting.

The same ablation also clarifies the role of the subdivision component.
Its effect is strongly length-dependent.
At $T{=}60$ on UCI, the raw profile already contains enough samples to
support stable estimation, and the subdivided version performs slightly
worse than Heis-noSH.
At $T{=}8$ on Pen Digits, the difference reverses but remains small on
clean data.
This pattern supports the interpretation that $S_H$ is not introduced to
improve already stable clean-data performance, but to provide a denser
and more robust estimate of the signed-area profile when sampling is
sparse or when noise is present.
That interpretation is consistent with the noise results in
Tables~\ref{tab:uci_noise} and~\ref{tab:pendigits_noise}, where the full
Euc+Heis descriptor exceeds Euc+$z(T)$ by a further $+4.2$\,pp on UCI
and $+5.0$\,pp on Pen Digits at $\sigma{=}0.20$.

A second important conclusion is that the gains are genuinely geometric.
The dimension-matched controls in Table~\ref{tab:ablation} and the
McNemar results in Table~\ref{tab:mcnemar} show that random feature
augmentation does not reproduce the effect.
Euc+Rand is not significantly different from Euclidean, whereas
Euc+$z(T)$ is highly significant on both datasets.
The same pattern holds at the fuller level of comparison: Euc+Rand$_{15}$ remains statistically indistinguishable from Euclidean, while Euc+Heis yields a
large and significant improvement.
This is an important result for the interpretation of the paper, because
it shows that the benefit does not arise from merely increasing the
descriptor dimension, but from introducing order-sensitive information
that Euclidean features do not encode.

\begin{table}[t]
\caption{McNemar significance tests (RF, 5-fold CV pooled predictions,
  corrected statistic, df$\,{=}\,1$).
  Euc+Rand (26-dim) and Euc+Rand$_{15}$ (40-dim) are dimension-matched
  random baselines for Euc+$z(T)$ and Euc+Heis respectively.}
\label{tab:mcnemar}
\centering\small
\begin{tabular}{llccl}
\toprule
Dataset & Comparison & $\chi^2$ & $p$ & Result \\
\midrule
UCI  & Euc vs Euc+Rand (26-dim)    &   0.00 & $1.000$  & n.s. \\
UCI  & Euc vs Euc+Rand$_{15}$ (40-dim) & 0.12 & $0.731$ & n.s. \\
UCI  & \textbf{Euc vs Euc+$z(T)$} & \textbf{105.33} & $<0.001$ & \textbf{sig.} \\
UCI  & Euc vs Euc+Heis            & 107.23 & $<0.001$ & sig. \\
UCI  & Euc+$z(T)$ vs Euc+Heis    &   6.73 & $0.009$  & sig. \\
Pen  & \textbf{Euc vs Euc+$z(T)$} & \textbf{163.99} & $<0.001$ & \textbf{sig.} \\
Pen  & Euc vs Euc+Heis            & 351.86 & $<0.001$ & sig. \\
Pen  & Euc+$z(T)$ vs Euc+Heis    & 131.60 & $<0.001$ & sig. \\
\bottomrule
\end{tabular}
\end{table}

From a document-analysis perspective, the results identify three
conditions under which the proposed descriptor is most useful.
In clean multi-class recognition of moderate-length handwriting
trajectories, appending $z(T)$ alone provides most of the available
benefit at negligible additional cost and without requiring any change
to the underlying pipeline.
In hard same-shape confusions of the kind that arise between characters
with similar spatial outlines but opposite traversal directions, the
signed area can itself become the decisive signal, as is demonstrated by
the \texttt{o} versus \texttt{y} pair where Euclidean features plateau
at $0.964$ while $z(T)$ alone achieves perfect separation.
In noisy pen-trajectory data, the full Heisenberg profile is the more
attractive option, since the accuracy advantage of Euc+Heis over the
Euclidean baseline grows substantially with increasing noise, reflecting
the greater stability of the subdivided profile under coordinate
perturbation.
These three conditions lead to a straightforward practical
recommendation for document-analysis practitioners: appending $z(T)$
suffices for a lightweight upgrade to an existing pipeline, while the
full Heisenberg descriptor is appropriate when the trajectories are
noisy, short, or strongly shaped by loop dynamics.

The interpretability of the signed-area profile is one of the more
useful aspects of the method.
Closed-loop characters such as \texttt{o}, \texttt{a}, and \texttt{e}
produce a large and largely monotone accumulation of signed area, whereas
open strokes such as \texttt{n}, \texttt{u}, and \texttt{y} generate
profiles that oscillate near zero because the trajectory doubles back
and cancels previously accumulated area.
This geometric distinction explains why the method succeeds on cases
where Euclidean shape descriptors fail. The shapes may be similar in
space, but the manner in which they are traced is different.
The summary statistics extracted from the profile then encode whether
this accumulation is monotone, oscillatory, concentrated early or late,
or subject to reversals.

These results should also be read alongside the stronger baselines in
the time-series classification literature and alongside the
path-signature framework.
ROCKET clearly outperforms the proposed descriptor on both handwriting
benchmarks.
This is expected: ROCKET employs a substantially larger and less
interpretable feature representation optimised for maximum predictive
accuracy.
The proposed descriptor occupies a different position on the tradeoff
between accuracy, interpretability, and computational cost.
It is compact, closed-form, parameter-free, and directly tied to a
geometric quantity with a clear and elementary meaning in the context of
online handwriting, namely the area swept by the pen stroke and the
direction in which it is swept.
The comparison with path signatures also requires careful framing.
The signed area is well-established as the antisymmetric level-two
signature term, so the contribution of the present work is not the
introduction of a new mathematical invariant.
Rather, it is the demonstration that this quantity can be packaged into
a compact, temporally resolved profile that yields consistent
improvements on established online handwriting benchmarks at low feature
dimension and negligible computational overhead.

It is also important to state the scope of the present study precisely.
The empirical evaluation is restricted to two established online
handwriting benchmarks, the UCI Character Trajectories dataset and the
UCI Pen Digits dataset, and robustness is assessed specifically under
additive Gaussian coordinate noise.
These choices reflect the document-analysis focus of the work and the
availability of writer-level metadata: writer-independent splits are
directly available for Pen Digits, while the Character Trajectories
dataset does not release the writer identifiers needed to construct
strict writer-independent partitions.
Other perturbation families relevant to real document-acquisition
pipelines, including point dropout, temporal resampling, and local
geometric distortion, remain natural targets for future investigation.
Extension of the Heisenberg-lift construction to higher-dimensional
trajectory spaces would also require a different nilpotent-group
formulation and lies beyond the scope of the present paper.
These boundaries do not diminish the central findings, but they define
the setting within which the conclusions are properly applicable, namely
online handwriting and related two-dimensional pen-trajectory
recognition tasks in document analysis.

%%=========================================================%%
\section{Conclusion}
\label{sec:conclusion}
%%=========================================================%%

This paper has addressed a specific shortcoming of conventional
Euclidean descriptors for online handwriting trajectories in document
analysis.
Standard shape features represent the spatial outline of a pen stroke
effectively, but they discard the order in which that outline is
produced.
When the discriminative information between character classes lies in
loop orientation, stroke direction, or traversal sequencing rather than
in spatial shape, this omission is not recoverable by any amount of
additional Euclidean information.
To fill this gap, we introduced a Heisenberg-lift framework that
extracts compact, interpretable, and order-sensitive features from
planar pen trajectories, and evaluated it on two standard online
handwriting benchmarks under clean and noise-degraded conditions.

The principal empirical finding is that even the minimal form of the
method provides consistent and statistically significant benefit.
Appending the terminal signed area $z(T)$ as a single additional
scalar to a conventional 25-dimensional Euclidean descriptor yields
improvements on both benchmarks, while introducing no learned
parameters and negligible computational overhead.
On the hardest character pair identified in the study, \texttt{o}
versus \texttt{y}, this single scalar achieves perfect separation,
whereas the 25-dimensional Euclidean baseline falls short of that
level.
The result makes the central point concretely: when the class
distinction is carried by traversal direction rather than by spatial
shape, an order-sensitive quantity recovers information that
shape-only descriptors cannot, regardless of their dimensionality.

The richer Heisenberg descriptor, derived by tracking the temporal
dynamics of signed-area accumulation through a compact profile of
summary statistics, extends the gains further.
It outperforms the minimal augmentation on clean data and yields
substantially larger improvements under additive noise, particularly
for short or loop-structured trajectories of the kind common in
handwritten character recognition.
Dimension-matched random controls and McNemar significance tests
confirm across both datasets that these improvements originate in the
geometric content of the descriptor rather than in feature-count
inflation.

The scope of the present study is deliberately focused.
The evaluation covers two established online handwriting benchmarks
and assesses robustness under additive Gaussian coordinate noise, which
is a practically relevant degradation in pen-trajectory data.
The signed area is already recognised in the path-signature literature
as the antisymmetric level-two term, and the contribution here is not
the introduction of a new invariant.
Rather, the contribution is its systematic packaging as a
temporally resolved, interpretable, low-dimensional profile that
integrates naturally into existing online handwriting and
document-trajectory classification pipelines.
Broader validation across additional handwriting corpora, alternative
noise and perturbation models, and other document-analysis settings
involving pen trajectories constitute natural directions for future
work.

The practical implication for document-analysis practitioners is
direct.
When traversal direction is expected to carry class-relevant
information, the terminal signed area can be appended to any existing
Euclidean pipeline as a single extra feature at essentially zero
additional cost.
Where trajectories are short, noisy, or structured by complex loop
dynamics, the full Heisenberg profile provides a stronger
alternative that remains lightweight, closed-form, and directly
interpretable in terms of the geometric content of each pen stroke.

\section*{Acknowledgments}

The authors acknowledge the computational resources provided by the
Centre for Visual Computing and Intelligent Systems at the University
of Bradford.

\section*{Data Availability}

The UCI Character Trajectories and UCI Pen-Based Digit Recognition
datasets are publicly available at the UCI Machine Learning Repository.
For replication of the results, the full implementation is provided as a Google Colab notebook and is available at: \\
    \url{https://github.com/ugail/signed_area_heisenberg_trajectory_classification}

\section*{Funding}
No funding was received for this work.

\section*{Competing interests}
The authors declare no competing interests.

\section*{Author contributions}
Both authors contributed equally to this work.

%=======================================================================

\end{document}